\documentclass[sigconf]{acmart}
\renewcommand\footnotetextcopyrightpermission[1]{} 
\usepackage{enumitem}
\usepackage{graphicx}
\usepackage{tikz}
\usepackage{soul}
\usetikzlibrary{shapes, arrows, positioning, fit, backgrounds, calc, shadows}
\usetikzlibrary{shapes.geometric, shapes.symbols, patterns, fit, backgrounds, calc , shapes.misc}
\usepackage{array}
\graphicspath{ {./figs/} }
\AtBeginDocument{%
  }

\copyrightyear{}
\acmYear{}
\acmDOI{}
\acmConference[MLBench '26]{Sixth Workshop on Benchmarking Machine Learning Workloads on Emerging Hardware}{March 22, 2026}{Pittsburgh, PA, USA}
\acmYear{2026}
\copyrightyear{2026}

\acmISBN{}

\begin{document}


\title{CTS-Bench: Benchmarking Graph Coarsening Trade-offs for GNNs in Clock Tree Synthesis}
\subtitle{Accepted to ML Bench'26 Workshop ASPLOS }

\author{Barsat Khadka}
\affiliation{%
  \institution{The University of Southern Mississippi}
  \city{Hattiesburg}
  \state{Mississippi}
  \country{USA}}
\email{Barsat.Khadka@usm.edu}

\author{Kawsher Roxy}
\affiliation{%
  \institution{Intel Corporation}
  \city{Hillsboro}
  \state{Oregon}
  \country{USA}}
\email{kawsher.roxy@intel.com}

\author{Md Rubel Ahmed}
\affiliation{%
  \institution{Louisiana Tech University}
  \city{Ruston}
   \state{Louisiana}
  \country{USA}}
\email{mahamed@latech.edu}


\begin{abstract}
Graph Neural Networks (GNNs) are increasingly explored for physical design analysis in Electronic Design Automation, particularly for modeling Clock Tree Synthesis behavior such as clock skew and buffering complexity. However, practical deployment remains limited due to the prohibitive memory and runtime cost of operating on raw gate-level netlists. Graph coarsening is commonly used to improve scalability, yet its impact on CTS-critical learning objectives is not well characterized.
This paper introduces CTS-Bench, a benchmark suite for systematically evaluating the trade-offs between graph coarsening, prediction accuracy, and computational efficiency in GNN-based CTS analysis. CTS-Bench consists of 4,860 converged physical design solutions spanning five architectures and provides paired raw gate-level and clustered graph representations derived from post-placement designs.
Using clock skew prediction as a representative CTS task, we demonstrate a clear accuracy–efficiency trade-off. While graph coarsening reduces GPU memory usage by up to $17.2\times$ and accelerates training by up to $3\times$, it also removes structural information essential for modeling clock distribution, frequently resulting in negative $R^2$ scores under zero-shot evaluation. Our findings indicate that generic graph clustering techniques can fundamentally compromise CTS learning objectives, even when global physical metrics remain unchanged.
CTS-Bench enables principled evaluation of CTS-aware graph coarsening strategies, supports benchmarking of GNN architectures and accelerators under realistic physical design constraints, and provides a foundation for developing learning-assisted CTS analysis and optimization techniques. Code publicly available at: https://github.com/BarsatKhadka/CTS-Bench
\end{abstract}



\maketitle

\section{Introduction}
\label{sec:motivation}

Clock Tree Synthesis (CTS) is one of the most structurally sensitive stages of physical design, where small variations in placement topology can lead to disproportionate changes in clock skew, buffering overhead, and power consumption. Unlike logic optimization, CTS operates under largely fixed placement and routing constraints, forcing the clock network to adapt to the spatial and hierarchical organization established earlier in the design flow. As the end of Dennard scaling has significantly increased design complexity, the EDA community has increasingly turned to machine learning techniques to improve scalability and efficiency across the RTL-to-GDSII pipeline \cite{huang2021survey}.

Among these techniques, GNNs have gained particular attention because the logical connectivity of digital circuits is inherently graph-structured. By directly modeling netlists and their structural dependencies, graph-based methods and GNNs have shown promise across multiple physical design tasks, including placement \cite{Mirhoseini2021GraphPlacement,lu2020vlsi,dinh2024graph}, routing \cite{cheng2021joint}, timing prediction \cite{Hu2023SyncTREE}, and Clock Tree Synthesis \cite{wu2025graph}. These approaches suggest that learning-based surrogates may capture complex structural interactions that are difficult to express using traditional analytical models.

However, the development, evaluation, and benchmarking of such models and their specialized accelerators remain severely constrained by the lack of public, standardized datasets, largely due to the proprietary nature of industrial semiconductor designs. While prior efforts have introduced graph-based EDA frameworks \cite{shrestha2024eda} and design phase specific benchmarks \cite{viswanathan2011ispd,jiang2024circuitnet}, existing datasets do not adequately capture the coupled interaction between placement-induced structure and CTS behavior. As a result, there is limited visibility into how structural abstractions and learning models impact CTS target metrics under realistic physical design conditions.

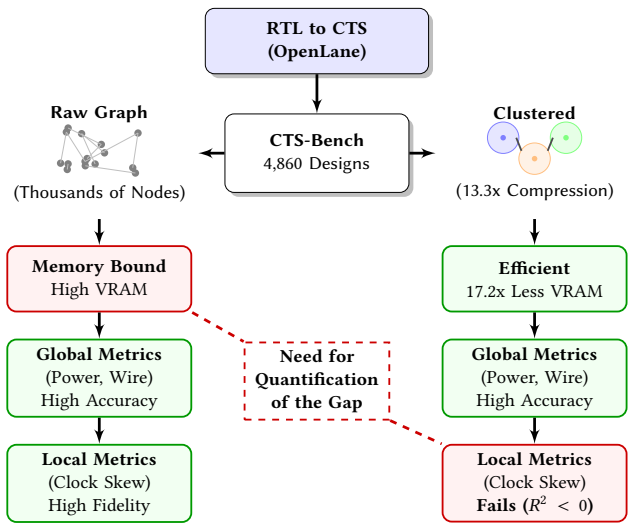
\begin{figure}[ht]
\centering
\resizebox{\columnwidth}{!}{%
\begin{tikzpicture}[
    auto,
    node distance = 0.6cm and 0.4cm,
    font=\small\sffamily,
    startstop/.style = {rectangle, draw, fill=blue!10, text width=3.2cm, text centered, rounded corners, minimum height=1.0cm, drop shadow, font=\bfseries\small},
    graphblock/.style = {rectangle, text width=2.8cm, text centered, minimum height=2.0cm},
    block/.style = {rectangle, draw, fill=white, text width=2.6cm, text centered, rounded corners, minimum height=1.2cm, drop shadow},
    line/.style = {draw, -latex', very thick},
    redbox/.style = {rectangle, draw=red!80!black, fill=red!5, thick, text width=2.6cm, align=center, rounded corners, minimum height=1.0cm, font=\small},
    greenbox/.style = {rectangle, draw=green!60!black, fill=green!5, thick, text width=2.6cm, align=center, rounded corners, minimum height=1.0cm, font=\small},
    gapbox/.style = {rectangle, draw=red!80!black, dashed, fill=white, thick, text width=2.0cm, align=center, font=\small\bfseries, inner sep=3pt}
]

    \node [startstop] (input) {RTL to CTS\\(OpenLane)};
    \node [block, below=0.6cm of input] (dataset) {\textbf{CTS-Bench}\\4,860 Designs};

    
    \node [graphblock, left=of dataset] (raw_graph) {
        \textbf{Raw Graph}\\
        \begin{tikzpicture}[scale=0.4, baseline=0]
            \foreach \i in {1,...,15} {
                \fill [gray] (rand*2.0, rand*1.0) coordinate (n\i) circle (4pt); 
            }
            \draw[gray!50, thin] (n1) -- (n4); \draw[gray!50, thin] (n2) -- (n7);
            \draw[gray!50, thin] (n3) -- (n10); \draw[gray!50, thin] (n5) -- (n9);
            \draw[gray!50, thin] (n6) -- (n8); \draw[gray!50, thin] (n1) -- (n12);
            \draw[gray!50, thin] (n13) -- (n2); \draw[gray!50, thin] (n14) -- (n15);
             \draw[gray!50, thin] (n3) -- (n5); \draw[gray!50, thin] (n9) -- (n11);
        \end{tikzpicture}\\[0.3em]
        (Thousands of Nodes)
    };
    
    \node [graphblock, right=of dataset] (cluster_graph) {
        \textbf{Clustered}\\
        \begin{tikzpicture}[scale=0.4, baseline=0]
            \draw[fill=blue!10, draw=blue!50] (-1.2, 0.4) circle (0.6);
            \fill [blue!50] (-1.2, 0.4) circle (3pt);
            
            \draw[fill=orange!10, draw=orange!50] (0, -0.4) circle (0.6);
            \fill [orange!50] (0, -0.4) circle (3pt);
            
            \draw[fill=green!10, draw=green!50] (1.2, 0.4) circle (0.6);
            \fill [green!50] (1.2, 0.4) circle (3pt);
            
            \draw[thick, black!70] (-0.7, 0.4) -- (-0.5, -0.1); 
            \draw[thick, black!70] (0.5, -0.1) -- (0.7, 0.4);
        \end{tikzpicture}\\[0.3em]
        (13.3x Compression)
    };

    \node [redbox, below=0.4cm of raw_graph] (raw_eff) {
        \textbf{Memory Bound}\\High VRAM
    };
    \node [greenbox, below=0.4cm of cluster_graph] (cluster_eff) {
        \textbf{Efficient}\\17.2x Less VRAM
    };

    \node [greenbox, below=0.4cm of raw_eff] (raw_global) {
        \textbf{Global Metrics}\\(Power, Wire)\\High Accuracy
    };
    \node [greenbox, below=0.4cm of cluster_eff] (cluster_global) {
        \textbf{Global Metrics}\\(Power, Wire)\\High Accuracy
    };

    \node [greenbox, below=0.4cm of raw_global] (raw_local) {
        \textbf{Local Metrics}\\(Clock Skew)\\High Fidelity
    };
    \node [redbox, below=0.4cm of cluster_global] (cluster_local) {
        \textbf{Local Metrics}\\(Clock Skew)\\\textbf{Fails ($R^2 < 0$)}
    };

    \path [line] (input) -- (dataset);
    \path [line] (dataset) -- (raw_graph);
    \path [line] (dataset) -- (cluster_graph);
    \path [line] (raw_graph) -- (raw_eff);
    \path [line] (raw_eff) -- (raw_global);
    \path [line] (raw_global) -- (raw_local);
    \path [line] (cluster_graph) -- (cluster_eff);
    \path [line] (cluster_eff) -- (cluster_global);
    \path [line] (cluster_global) -- (cluster_local);

    
    \coordinate (top_center) at ($(raw_eff.east)!0.5!(cluster_eff.west)$);
    \coordinate (bot_center) at ($(raw_local.east)!0.5!(cluster_local.west)$);

    \node [gapbox] (mid_box) at ($(top_center)!0.5!(bot_center)$) {Need for\\Quantification\\ of the Gap};

    \draw [very thick, dashed, red!80!black] (raw_eff.south east) -- (mid_box.north west);
    
    \draw [very thick, dashed, red!80!black] (mid_box.south east) -- (cluster_local.north west);

\end{tikzpicture}
}
\caption{ CTS-Bench benchmarks the trade-off between memory efficiency and model fidelity }
\label{fig:marketing_vertical}
\end{figure}
Graph learning has emerged as a promising approach for CTS analysis, but it introduces extreme computational demands that challenge current systems. Modeling designs at the gate level results in graphs with millions of nodes, making direct learning impractical. Graph clustering provides a natural mechanism to improve scalability, but its impact on CTS-relevant learning fidelity and compute efficiency is not well quantified. This gap motivates the need for benchmarks that explicitly characterize the trade-offs between graph resolution and computational cost.

 CTS-Bench provides paired raw gate-level graphs, which stress memory capacity, and physics-aware clustered graphs, which enable scalable learning and efficiency analysis. These representations are paired with ground-truth CTS metrics obtained from converged physical design flows. Beyond model accuracy, CTS-Bench enables systematic evaluation of system-level costs, allowing researchers to study how graph reduction and physics-informed clustering affect memory usage, runtime, and learning efficiency on standard hardware. Figure~\ref{fig:marketing_vertical} shows the existing gap in graph clustering approaches. Our primary contributions are fourfold:
\begin{itemize}[noitemsep, topsep=2pt, partopsep=0pt]
\item  Solution Space: We collect 4,860 data points across five logic architectures. For each placement, we generate 10 CTS variants, capturing the relationship between placement decisions and CTS outcomes. This provides rich data for benchmarking GNN scalability on real workloads
\item Multi-Scale Representations: We provide both raw and clustered proxy graphs to create a \textit{scalability benchmark}. This evaluates the trade-off between hardware efficiency (memory usage) and model accuracy, determining if compressed topologies can serve as viable proxies.
\item \texttt{Gap} Quantification: We introduce scoring metrics to evaluate the constructibility of a clock tree over a given placement. This measures the gap between the placement's structural quality and the final CTS target values across three hard constraints: ( Skew, Power, and Wirelength ).
\item Reproducible Generation Framework: We release our automated, containerized data generation framework built on OpenLane. This allows researchers to extend the benchmark to new architectures or technology nodes, for a continuous and reproducible ecosystem for ML-EDA research.
\end{itemize}
The remainder of this paper is organized as follows. Section~\ref{sec:literature} reviews related benchmarking efforts in physical design. Section~\ref{sec:methodology} describes the benchmark data generation process. Section~\ref{sec:experiments} presents the experimental setup and evaluated workloads. Section~\ref{sec:result} analyzes the benchmarking results, and Section~\ref{sec:conlcude} concludes the paper with a discussion of future directions.

\medskip
\noindent \textbf{Related Works.}
\label{sec:literature}
The evolution of physical design evaluation has shifted from purely algorithmic competitions to data-driven predictive modeling. The following paragraphs contextualizes CTS-Bench within the landscape of traditional benchmarks, standardized graph schemas, and recent learning-centric datasets.

\noindent
\textbf{Traditional Physical Design Benchmarks:} There has been continued interest in developing benchmark suites for Physical Design evaluation. Classical contest benchmark suites such as ISPD98~\cite{10.1145/274535.274546}, ISPD2005~\cite{10.1145/1055137.1055182}, and ICCAD2015~\cite{7372671}  have long served as standards for evaluating physical design algorithms. While valuable, they do not provide explicit rules to construct graph-structured representations suitable for GNN-based learning and benchmarking. Each instance also typically has a single reference solution, limiting the study of design sensitivity. Modern data-driven models benefit from a dense solution space that captures variations in design quality, enabling learning across multiple configurations.

\noindent
\textbf{General Graph Learning Benchmarks:} The broader ML community has established rigorous benchmarks such as the Open Graph Benchmark (OGB) \cite{hu2021opengraphbenchmarkdatasets} and Long-Range Graph Benchmark (LRGB) \cite{dwivedi2023longrangegraphbenchmark}. While these have standardized evaluation for social, citation, and molecular graphs, they do not reflect the unique characteristics of EDA workloads. Logic netlists exhibit distinct properties: massive scale, high-fanout clock nets, and strict directed signal flows that are absent in standard datasets. CTS-Bench bridges this gap by providing a domain-specific challenge that is structurally distinct from general-purpose graph benchmarks.

\noindent
\textbf{Graph-Based Schema for EDA:} Recently, EDA-schema \cite{10.1145/3649476.3658718} introduced a standardized graph format for representing circuits across the entire physical design flow. While it provides a standardized infrastructure for general-purpose design data, its initial dataset of 960 designs focuses on broad PPA baseline estimation across the full RTL-to-GDSII flow. 

However, these approaches do not isolate the specific, coupled interaction at the Placement-CTS interface. CTS-Bench addresses this gap by providing a specialized benchmark for this stage.

\section{CTS-Bench Methodology}
\label{sec:methodology}

The CTS-Bench pipeline is built to systematically explore the placement and CTS relationship by using OpenLane, OpenROAD, TritonCTS with Sky130 PDK, and offers two different graph scales per placement to benchmark how model performance changes under varying hardware memory constraints. All results, including 15 Quality-of-Result (QoR) metrics and the paths to their corresponding graph files, are stored in a central manifest for consistent benchmarking.

\subsection{Automated Placement Generation Pipeline}

We employ an open-source RTL-to-GDSII flow using OpenROAD\cite{Ajayi2019OpenROADTA} and the SkyWater 130nm PDK, containerized in nix-shell to ensure reproducibility across different compute platforms. Our dataset includes five hardware designs with diverse computational patterns: PicoRV32, AES, SHA256, and EthMAC for training, and an additional 500 data points from a Zipdiv core for zero-shot generalization testing. The generation depicted in Figure~\ref{fig:pipeline} comprises of three stages:

\noindent\textbf{Stage 1: Placement Generation and Activity Extraction.} Starting from RTL code, we execute synthesis, floorplanning, power network insertion, and detailed placement, stopping before CTS. To create workload diversity, we randomize seven placement parameters referenced in Table~\ref{tab:knobs}. These parameters fundamentally alter spatial cell distribution, especially with a varying synthesis strategy. We generate 486 unique placements across all five designs. We also perform gate-level simulation using design-specific testbenches to capture switching activity (SAIF) for each unique placement.

\begin{table}[h]
\caption{Randomization Knobs for Data Diversity}
\label{tab:knobs}
\centering
\small
\begin{tabular}{lllc}
\toprule
\textbf{Stage} & \textbf{Parameter} & \textbf{Range / Set} & \textbf{Type} \\
\midrule
Synthesis & Synth Strategy & \{AREA 0-2, DELAY 0-4\} & Categorical \\
Floorplan & Aspect Ratio & \{0.7, 1.0, 1.4, 2.0\} & Discrete \\
Floorplan & IO Mode & \{Random Pin, Standard\} & Binary (0/1) \\
Placement & Core Utilization & 40\% -- 70\% & Continuous \\
Placement & Target Density & $Util + [0.0, 0.20]$ & Continuous \\
Placement & Time Driven & \{Enabled, Disabled\} & Binary (0/1) \\
Placement & Routability Driven & \{Enabled, Disabled\} & Binary (0/1) \\
\midrule
CTS & Sink Max Dia. & 35 -- 70 $\mu m$ & Integer \\
CTS & Max Wire Length & 130 -- 280 $\mu m$ & Integer \\
CTS & Cluster Size & 12 -- 30 sinks & Integer \\
CTS & Buffer Distance & 70 -- 150 $\mu m$ & Integer \\
\bottomrule
\end{tabular}
\end{table}

\begin{table}[h]
\caption{Labels used for QoR Metrics}
\label{tab:metrics}
\centering
\small
\begin{tabular}{llc}
\toprule
\textbf{Category} & \textbf{Metric} & \textbf{Unit} \\
\midrule
\textbf{Timing} & Skew (Setup/Hold), Slack, TNS & $ns$ \\
\textbf{Power} & Total Dynamic/Static Power & $W$ \\
\textbf{Physical} & Wirelength, Cell Utilization & $\mu m, \%$ \\
\textbf{Resources} & Clock Buffers/Inverters Count & integer \\
\textbf{Congestion} & Routing/Repair Buffer Count & integer \\
\bottomrule
\end{tabular}
\end{table}

\noindent \textbf{Stage 2: Multi-Scale Graph Construction.}
To evaluate the trade-off between model fidelity and hardware efficiency, we transform each physical placement into two graph representations: a fine-grained Raw Graph and a physics-aware Clustered Graph.

\noindent\textbf{(a) Raw Graph Construction:} Nodes represent standard cells (flip-flops and combinational logic), and edges represent logical connectivity. We prioritize flip-flops and their one-hop fan-out logic, as data-path cells immediately surrounding registers have the strongest impact on clock tree behavior\cite{10.1145/3670474.3685949}. Node features include: (1) normalized geometric coordinates (x,y), (2) a binary cell-type indicator (is\_ff), and (3) log-transformed switching activity from SAIF data. Edges are weighted by the Manhattan distance between connected cells: $d_\text{Manhattan}(c_i, c_j) = |x_i - x_j| + |y_i - y_j|$.
\pgfdeclarelayer{background}
\pgfdeclarelayer{foreground}
\pgfsetlayers{background,main,foreground}

\begin{figure}[ht]
\centering
\resizebox{\columnwidth}{!}{%
\begin{tikzpicture}[
    node distance=0.4cm and 0.4cm,
    font=\scriptsize\sffamily,
    process/.style = {
        rectangle, draw=black!70, thick, fill=white,
        text width=3.2cm,
        align=center, rounded corners,
        minimum height=0.9cm, drop shadow,
        font=\scriptsize\sffamily
    },
    visualstep/.style = {
        rectangle, draw=none, fill=none,
        text width=3.6cm,
        inner sep=2pt
    },
    container/.style = {
        rectangle, draw=gray!80, dashed, thick, fill=gray!5,
        inner sep=0.3cm, rounded corners
    },
    processbox/.style = {
        rectangle, draw=black, thick, fill=white,
        inner sep=0.25cm,
        rounded corners, drop shadow
    },
    arrow/.style = {-latex, thick, draw=black!80, rounded corners=3pt},
    imgnode/.style = {inner sep=0pt, outer sep=0pt},
    widefile/.style = {
        draw=black!60, thick, fill=white, shape=rectangle,
        text width=3.0cm, minimum height=1.4cm,
        align=center, font=\scriptsize\sffamily,
        append after command={
            \pgfextra{
                \draw[black!60] (\tikzlastnode.north east) -- ++(-0.2,0) -- ++(0,-0.2) -- cycle;
            }
        }
    }
]

    \node[process] (placement) {
        \textbf{Placement Generation}\\
        (OpenROAD)\\
        \textit{Rand. Knobs}
    };

    \node[process, below=0.3cm of placement] (activity) {
        \textbf{Activity Extraction}\\
        Icarus Verilog $\to$ VCD $\to$ SAIF
    };

    \begin{pgfonlayer}{background}
        \node[container, fit=(placement) (activity)] (stage1_box) {};
    \end{pgfonlayer}
    \node[anchor=north east, font=\bfseries\tiny, color=gray!80] at (stage1_box.north east) {Stage 1};

    \node[imgnode, left=0.6cm of stage1_box.west] (rtl_icon) {\includegraphics[width=0.7cm]{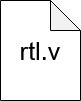}};
    \node[above=0.0cm of rtl_icon, font=\bfseries\tiny] {RTL-Design};

    \node[imgnode, right=0.6cm of stage1_box.east] (tb_icon) {\includegraphics[width=0.7cm]{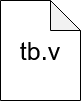}};
    \node[above=0.0cm of tb_icon, font=\bfseries\tiny] {TB};

    \draw[arrow] (rtl_icon.east) -- (stage1_box.west);
    \draw[arrow] (tb_icon.west) -- (stage1_box.east);
    \draw[arrow] (placement) -- (activity);

    \coordinate (split_point) at ($(stage1_box.south) + (0,-0.4)$);
    \draw[thick] (stage1_box.south) -- (split_point);


    \node[visualstep] (raw) at ($(split_point) + (-2.2, -1.2)$) {
        \begin{tabular}{@{}m{1.2cm} m{2.2cm}@{}}
            \textbf{Raw Graph} &
            \begin{tikzpicture}[scale=0.25, baseline=0]
                \foreach \i in {1,...,25} {
                    \fill [gray!70] (rand*1.8, rand*0.8) coordinate (n\i) circle (3pt);
                }
                \foreach \i in {1,...,15} {
                    \draw[gray!30, thin] (n\i) -- (n\the\numexpr\i+1\relax);
                }
            \end{tikzpicture}
        \end{tabular}
    };


    \node[anchor=center, font=\bfseries\footnotesize, below=0.65cm of raw] (proc_title) {Processing};

    \node[visualstep, below=0.1cm of proc_title] (step1) {
        \begin{tabular}{@{}m{1.2cm} m{2.2cm}@{}}
            \textbf{I. Atomic BFS} &
            \begin{tikzpicture}[scale=0.25, baseline=-3pt]
                \draw[fill=blue!20, thick] (0.5, 0.5) circle (0.5);
                \draw[-latex, thin] (1.1, 0.5) -- (2.0, 0.5);
                \draw[fill=white] (2.4, 0.5) circle (0.3);
                \draw[fill=white] (3.2, 0.8) circle (0.3);
                \draw[dashed, red!80, rounded corners=2pt] (-0.2, -0.2) rectangle (3.6, 1.2);
            \end{tikzpicture}
        \end{tabular}
    };

    \node[visualstep, below=0.0cm of step1] (step2) {
        \begin{tabular}{@{}m{1.2cm} m{2.2cm}@{}}
            \textbf{II. Spread Filter} &
            \begin{tikzpicture}[scale=0.25, baseline=-3pt]
                \draw[dotted] (0,0) circle (0.8); 
                \node[red, font=\bfseries] at (0,0) {$\times$};
                \node[font=\tiny] at (0, -1.4) {$\sigma\!>\!0.05$};
                
                \draw (1.8, -1) -- (1.8, 1);
                
                \draw[dotted] (3.5, 0) circle (0.5); 
                \node[green!60!black, font=\bfseries] at (3.5, 0) {$\checkmark$};
                \node[font=\tiny] at (3.5, -1.4) {Keep};
            \end{tikzpicture}
        \end{tabular}
    };

    \node[visualstep, below=0.0cm of step2] (step3) {
        \begin{tabular}{@{}m{1.2cm} m{2.2cm}@{}}
            \textbf{III. Gravity Merge} &
            \begin{tikzpicture}[scale=0.25, baseline=-3pt]
                \draw[fill=blue!10] (0,0) circle (0.3); \draw[-latex, blue] (0,0) -- (0.5, 0.5);
                \draw[fill=blue!10] (1,0) circle (0.3); \draw[-latex, blue] (1,0) -- (1.5, 0.5);
                \draw[-latex, thick] (1.8, 0.2) -- (2.4, 0.2);
                \draw[dashed, fill=green!10] (3.0, 0.2) circle (0.5);
            \end{tikzpicture}
        \end{tabular}
    };

    \node[visualstep, below=0.8cm of step3] (cluster) {
        \begin{tabular}{@{}m{1.2cm} m{2.2cm}@{}}
            \textbf{Clustered Graph} &
            \begin{tikzpicture}[scale=0.25, baseline=0]
                \draw[fill=blue!10, draw=blue!80, thick] (-1.2, 0.5) circle (0.7);
                \draw[fill=orange!10, draw=orange!80, thick] (0, -0.6) circle (0.7);
                \draw[fill=green!10, draw=green!80, thick] (1.2, 0.5) circle (0.7);
                \draw[thick, black!70] (-0.5, 0.5) -- (0.5, 0.5);
                \draw[thick, black!70] (-0.4, -0.2) -- (0.4, 0.2);
            \end{tikzpicture}
        \end{tabular}
    };

    \coordinate[below=0.4cm of cluster] (graph_pad);

    \begin{pgfonlayer}{background}
        \node[processbox, fit=(proc_title) (step1) (step2) (step3)] (proc_box) {};

        \node[container, fit=(raw) (cluster) (proc_box) (graph_pad), inner sep=0.2cm, yshift=0.25cm] (graph_box) {};
    \end{pgfonlayer}

    \node[anchor=north, font=\bfseries\footnotesize] at (graph_box.north) {Graph Construction};

    \draw[arrow] (split_point) -| (graph_box.north);
    \draw[-latex, thick, black!80] (raw.south) -- (proc_box.north);
    \draw[-latex, thick, black!80] (proc_box.south) -- (cluster.north);


    \node[process, text width=3.0cm] (cts) at ($(split_point) + (2.5, -1.2)$) {
        \textbf{Multi-Variant CTS}\\
        (TritonCTS)\\
        \textit{10x per Design}
    };

    \node[widefile, below=1.5cm of cts] (metadata) {
        \textbf{metadata.csv}\\
        Skew, Power, Wire\\
        Place \& CTS Knobs\\
        Gap Scores
    };

    \begin{pgfonlayer}{background}
        \node[container, fit=(cts) (metadata)] (cts_box) {};
    \end{pgfonlayer}

    \draw[arrow] (split_point) -| (cts_box.north);
    \draw[arrow] (cts) -- (metadata);

    \path (current bounding box.north west) ++(-0.2,0.2) coordinate (nw_pad);
    \path (current bounding box.south east) ++(0.2,-0.2) coordinate (se_pad);
    \useasboundingbox (nw_pad) rectangle (se_pad);

\end{tikzpicture}
}
\caption{CTS-Bench Data generation pipeline.}
\label{fig:pipeline}
\end{figure}
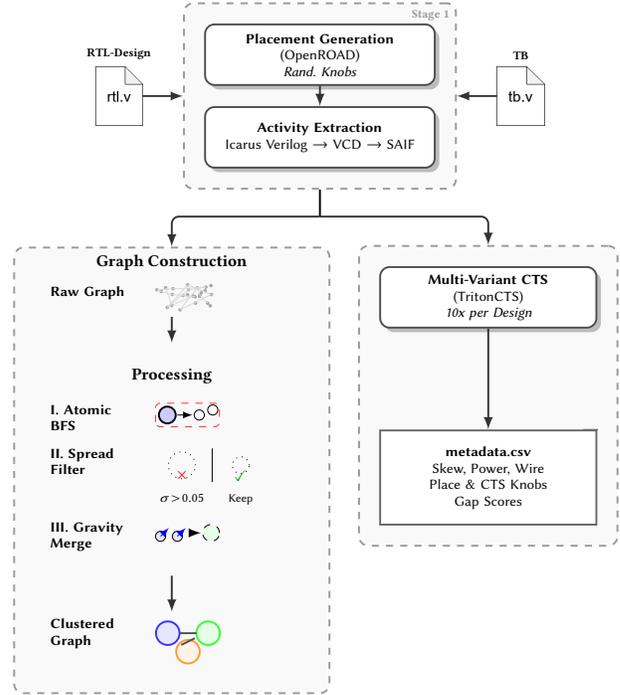

\begin{figure}[ht]
    \centering
    \includegraphics[width=0.8\columnwidth]{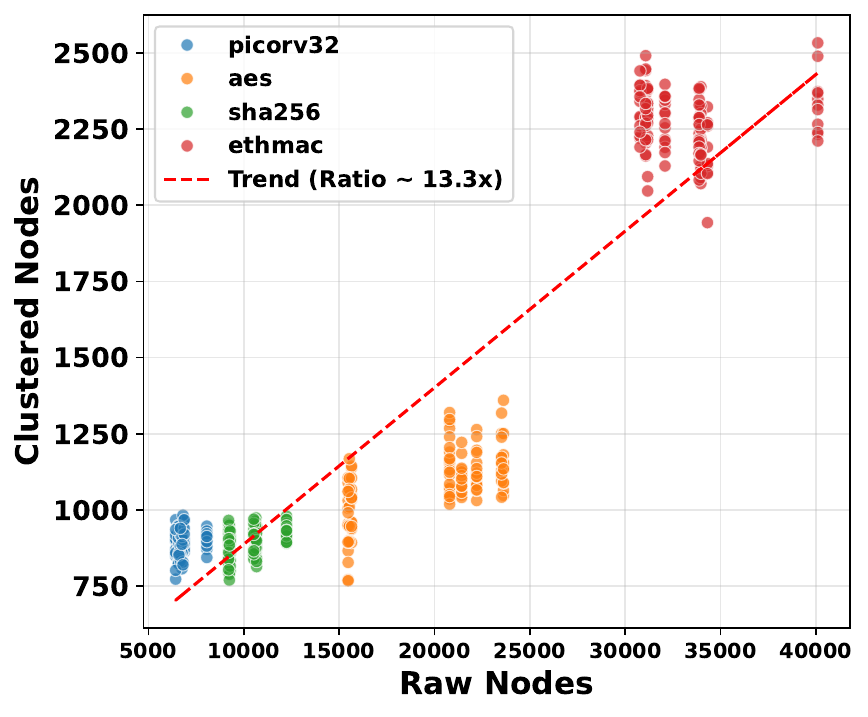}
    \caption{Raw Nodes are nodes in ground truth graphs, and Clustered nodes are nodes after coarsening. Vertical variance shows sensitivity to placement-specific
logic distribution. }
    \label{fig:compression}
\end{figure}

\noindent\textbf{(b) Clustered Graph Construction:} Raw graphs present significant memory overhead for GNN training. We develop a three-step clustering algorithm that achieves an approx 13.3$\times$ compression ratio while preserving the underlying representation of the design. Figure~\ref{fig:compression} shows graph compression for designs. 

\noindent\textbf{Step I. Atomic Cluster Formation:} We perform a breadth-first search (BFS) starting from each flip-flop. Each flip-flop claims its immediate combinational fan-out cone until another flip-flop or an already-claimed gate is encountered. This ensures every logic gate belongs to exactly one cluster. The flip flops are shuffled by randomizing so that there is no bias that everytime same flop claims the gate.

\noindent\textbf{Step II. High-Spread Filtering:} We identify high-spread atomic clusters by examining the spatial spread (standard deviation) of their member cells. Clusters with a spread exceeding 0.05 bypass further merging to preserve their unique geometric signature.

\noindent\textbf{Step III. Gravity-Vector-Aligned Merging:} For compact clusters, we perform controlled merging inspired by the concept of Multi-Bit Flip-Flops (MBFF). This stage aggregates atomic clusters into macro-nodes based on a few constraints: Merging is restricted to clusters sharing the same control net (e.g., reset or enable). This ensures the graph respects the design's underlying electrical domains. Gravity Vector: For each flip-flop, we calculate a Gravity Vector: a displacement vector pointing from the flip-flop’s position to the average coordinate (centroid) of its one-hop neighbors. This vector represents the gravitational pull that the surrounding logic exerts on that flip-flop.

We only merge clusters that share the same electrical compatibility, are physically close (Manhattan distance <0.05), and whose gravity vectors are highly aligned (cosine similarity >0.9). This ensures clustered nodes represent units that naturally pull in a common direction and share the same features. The final nodes contain 10 aggregated features: (1-2) geometric centroid $(x, y)$, (3-4) spatial spread $(\sigma_x, \sigma_y)$, (5) log-transformed cluster size, (6-7) composition counts ($N_{FF}$ vs. $N_{Logic}$), and (8-10) aggregated switching activity metrics (log-transformed max, sum, and non-zero toggle counts). Edge features are recomputed as the Manhattan distance between macro-cluster centroids, preserving the connectivity of the original atomic edges.

\vspace{0.5em}
\noindent \textbf{Stage 3: Multi-Variant CTS.} For each of the 486 placements, we generate 10 distinct clock trees by randomizing four TritonCTS parameters (Table 1). Each CTS run executes static timing analysis (STA) to extract 15 ground-truth metrics (Table 2). This yields 4,860 data points (486 placements $\times$ 10 CTS variants), capturing how different CTS policies respond to the same placement topology. All graphs are serialized as PyTorch Geometric tensors. Metadata includes CTS hyperparameters (inputs) and 15 metrics (targets) and respective graph paths in CSV format.

\subsection{Gap Calculation and Pareto Frontier}
To evaluate the placement-CTS interactions across heterogeneous architectures, we define a normalized \textit{Pareto Gap} framework. Raw metrics such as Skew (ps), Power (W), and Wirelength (units) are not directly comparable across designs of different scales. We therefore normalize each run $i$ of a specific design $D$ against the empirical best-case performance observed for that architecture.
For every run $i$ in a design $D$, we define the Gap Vector $\vec{G}_i$ as:
\begin{equation}
\vec{G}_i = \left[ \frac{Skew_i}{\min(Skew)_D}, \frac{Power_i}{\min(Power)_D}, \frac{WL_i}{\min(WL)_D} \right]
\end{equation}
 A run that achieves $\vec{G} = [1, 1, 1]$ represents the the point where all three hardware constraints are simultaneously minimized. To provide a singular scalar for model evaluation and ranking, we calculate the Total Pareto Distance ($D_{Pareto}$), representing the Euclidean distance from the ideal anchor
\begin{equation}
D_{Pareto} = \sqrt{(G_{Skew} - 1)^2 + (G_{Power} - 1)^2 + (G_{WL} - 1)^2}.
\end{equation}
This methodology enables CTS-Bench to characterize the overhead imposed by a placement configuration on the clock tree. For instance, a high $G_{Power}$ coupled with a low $G_{Skew}$ reveals a placement that requires aggressive buffer insertion to satisfy timing constraints, revealing the dominant bottleneck.

\section{Workload Characterization}
\label{sec:experiments}

We evaluate GNN-based EDA surrogates by comparing metrics of raw and clustered graph representations. Three GNN models: GCN \cite{kipf2017semisupervisedclassificationgraphconvolutional}, GraphSAGE \cite{hamilton2018inductiverepresentationlearninglarge}, and GATv2\cite{brody2022attentivegraphattentionnetworks} are trained for multi-target regression on post CTS metrics: clock skew, total power, and wirelength. We measure peak GPU memory (VRAM), training throughput, and performance metrics such as Mean Average Error (MAE) to analyze the trade-off between graph resolution, efficiency, and hardware usage. The experimental setup goes as follows:

\medskip
\noindent
\textbf{Multi-Modal Fusion Architecture:} 
We employ a Multi-Modal Fusion Architecture that integrates graph-based placement features with scalar placement and CTS tool knobs into a shared latent space. Graph embeddings are extracted using a GNN backbone (GCN, GraphSAGE, or GATv2) and pooled via global mean pooling. Placement and CTS knobs are processed through separate MLPs with different dimensions and learning rates as shown in Table~\ref{tab:params}, before being concatenated with the graph embedding. The fused representation is then passed through a fully connected head to predict three post-CTS metrics.

\noindent
\textbf{Dataset Splitting and Evaluation Strategy}
To test the model's ability and accuracy to generalize, we divide the data into Seen and Unseen categories:

 \textbf{1. In-Distribution (Seen):} The primary dataset consists of four diverse architectures (PicoRV32, AES, SHA256, EthMAC). We split this data randomly into 80\% Training and 20\% Validation sets. This evaluates how well the model performs on design types it has explicitly studied.

\textbf{2. Out-of-Distribution (Unseen):} To test if the model can adapt to completely new designs (zero-shot transfer), we hold out the Zipdiv architecture entirely. This set contains 500 data points that are excluded from the training process. Performance on this set is our primary metric for determining if the model can successfully handle unknown architectures.

Experiments were conducted on an NVIDIA GeForce RTX 5060 GPU 8GB VRAM (Driver 573.24), utilizing CUDA 12.8 for 100 epochs (for each representation per GNN) with a batch size of 32

\begin{table}[h]
\centering
\caption{Model Hyper-parameters used in CTS-Bench}
\vspace{-5pt}
\label{tab:params}
\begin{tabular}{lcccc}
\hline
\textbf{Graph} & \textbf{Hid. Dim.} & \textbf{L. Rate} & \textbf{Plac. Dim.} & \textbf{CTS Dim.} \\ \hline
Raw               & 64                  & 0.001                  & 32                     & 16              \\
Clustered         & 16                  & 0.0005                 & 8                      & 4               \\ \hline
\end{tabular}
\vspace{-5pt}
\end{table}

\section{Results and Discussion}
\label{sec:result}

We evaluate three standard GNN backbones: GCN, GraphSAGE, and GATv2 across both Raw and Clustered representations. Our analysis focuses on three dimensions: system efficiency, predictive fidelity, and zero-shot generalization.

\subsection{Memory and Compute Efficiency} 
\begin{figure}[t] 
    \centering
    \includegraphics[width=0.8\columnwidth]{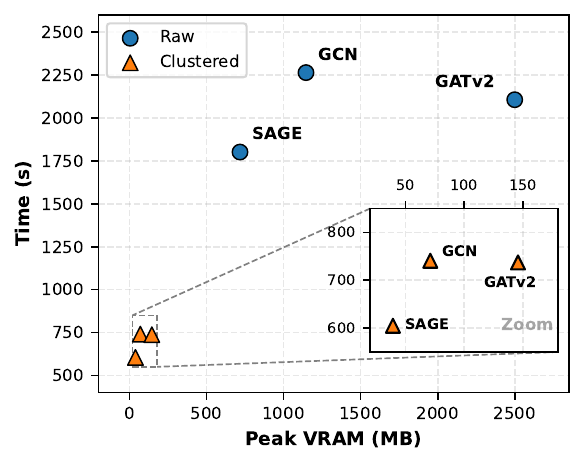}
    \caption{Efficiency benchmark comparing Raw and Clustered models in terms of Peak VRAM and execution time.}
    \label{fig:efficiency_benchmark}
\end{figure}

Figure~\ref{fig:efficiency_benchmark} highlights the computational bottleneck inherent in raw gate-level learning. Training a GATv2 model on raw graphs requires 1.2--2.5 GB of VRAM per batch, severely restricting throughput even on the moderate-scale partitions (5,000--40,000 nodes) used in this study. Scaling this approach to full industrial designs would likely exceed the memory capacity of standard accelerators. 

In contrast, the Clustered Graph representation effectively shifts the workload from memory-bound to compute-bound. By reducing the average node count by $\sim$13.3$\times$ across designs, we achieve an average of 17.2$\times$ reduction in VRAM usage and a 3$\times$ acceleration in training time. This confirms that clustering is a viable strategy on resource-constrained hardware.

\subsection{Accuracy Trade-off: Global vs. Local Metrics} \label{sec:accuracy} 

While clustering improves efficiency, it introduces a trade-off in model fidelity, particularly for local metrics. In this section, we explicitly analyze performance on seen architectures.

\begin{figure}[t] 
    \centering
    \includegraphics[width=1.0\columnwidth]{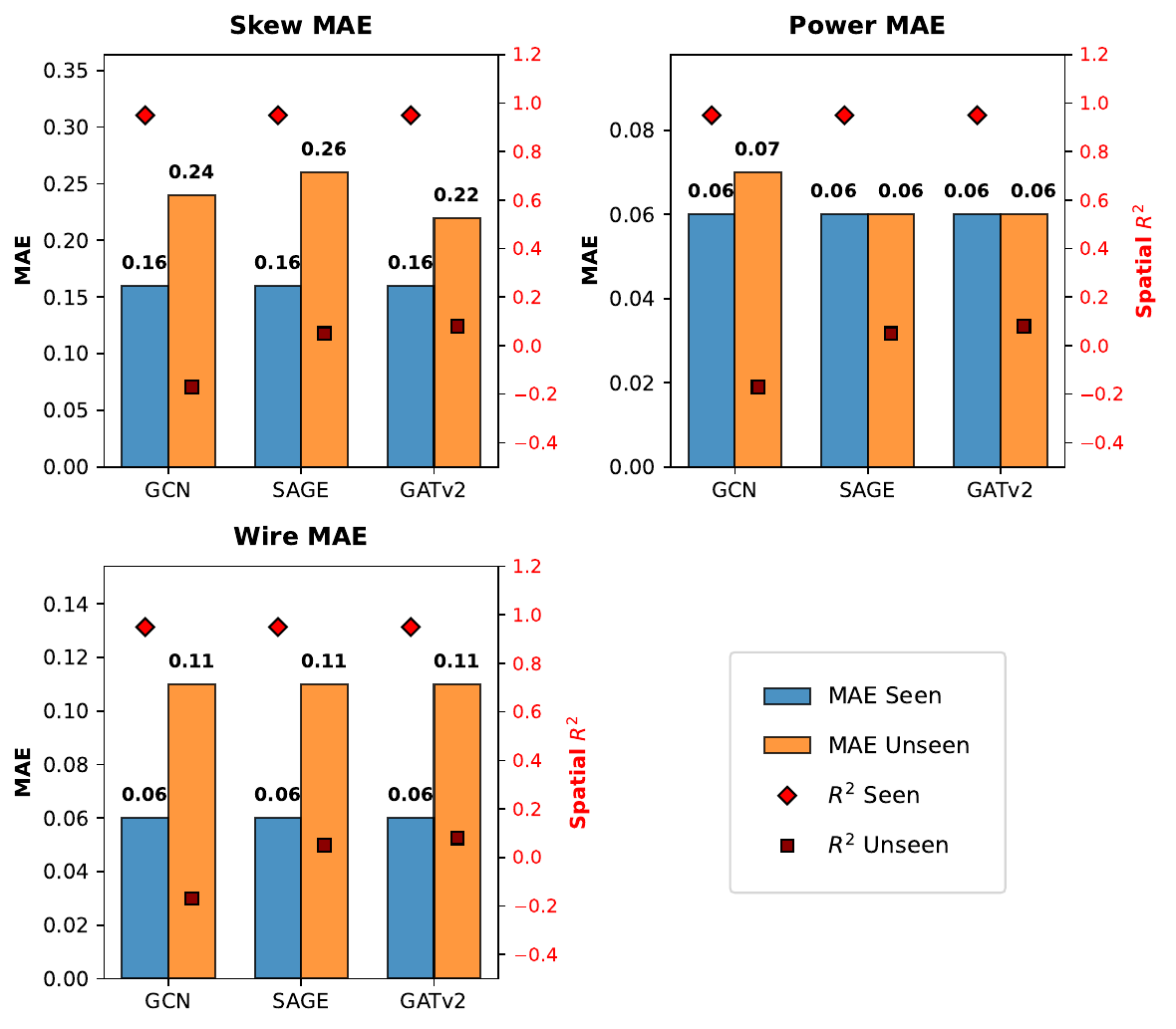}
    \caption{MAE Accuracy (bars, left y-axis) and Spatial $R^2$ Fidelity (markers, right y-axis) across Skew, Power, and Wire metrics for the Raw Workload.}
    \label{fig:raw_mae_r2}
\end{figure}

\begin{figure}[h!] 
    \centering
    \includegraphics[width=1.0\columnwidth]{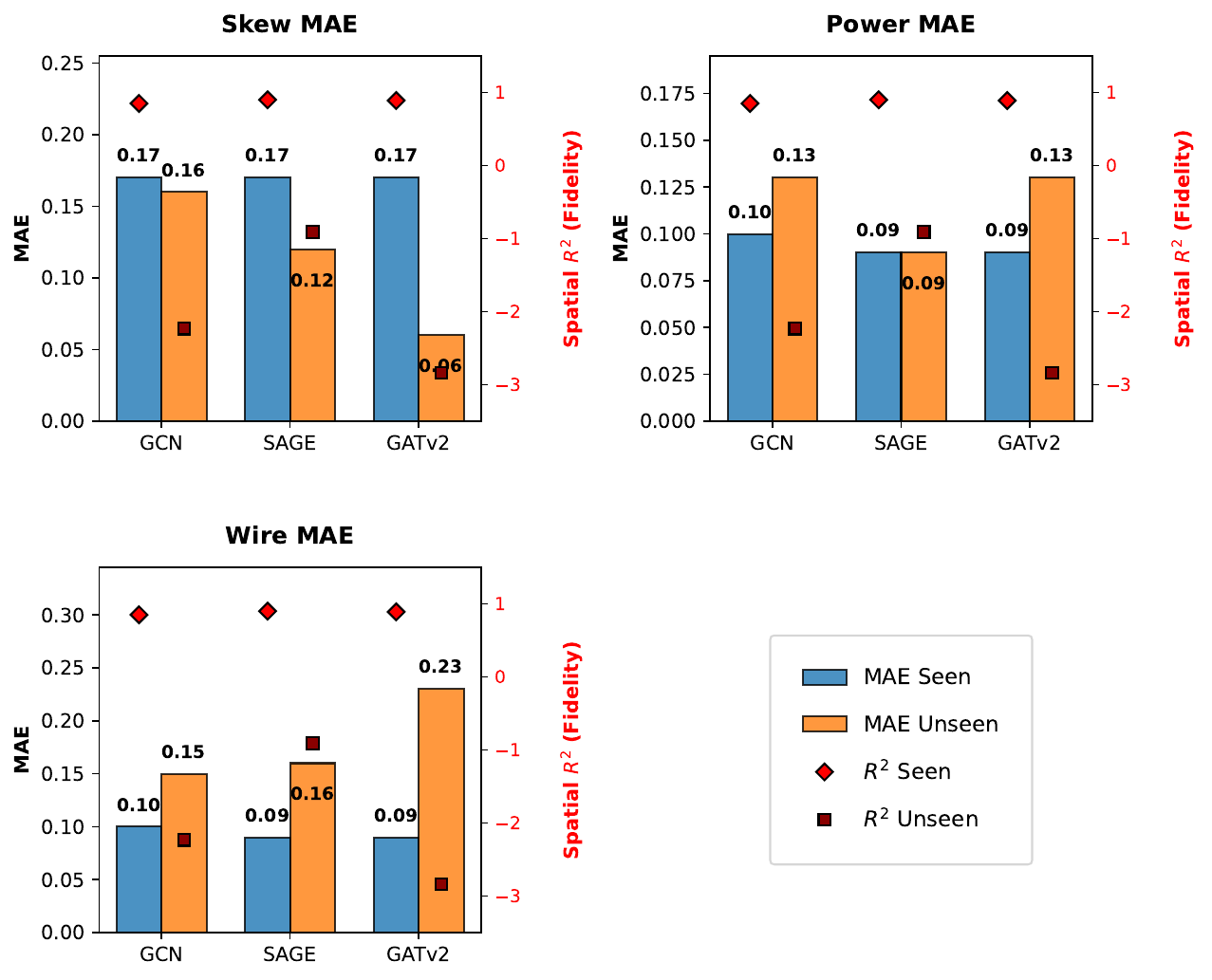}
    \caption{MAE Accuracy (bars, left axis) and Spatial $R^2$ Fidelity (markers, right axis) across Skew, Power, and Wire metrics for the Clustered Workload.}
    \label{fig:clustered_mae_r2}
\end{figure}

 \textbf{Global Metrics:} Global aggregates like Total Power and Wirelength remain highly predictable even under compression. As shown in Figure \ref{fig:clustered_mae_r2}, the Clustered GCN achieves a Power MAE of $0.10$ and Wirelength MAE of $0.10$ on seen data, which is comparable to the Raw baseline as shown in Figure \ref{fig:raw_mae_r2}, which was $0.06$ for both. This suggests that for seen designs, clustered proxies serve as a viable, high-efficiency alternative to raw gate-level graphs when dealing with global architectural features like wirelength and power

 \textbf{Local Metrics:} The abstraction cost is most visible in Clock Skew, a highly localized metric dependent on specific flop-to-flop distances. The Raw models achieve a Skew MAE of $0.16$ (GCN), whereas Clustered models degrade to $0.17$. While the MAE shift appears minor, the impact is evident in $R^2$. The degradation in $R^2$ indicates that clustered models lose the fine-grained resolution required to distinguish between high-skew and low-skew regions. 

 \textbf{The Fidelity and Robustness Gap:} A critical finding is that $R^2$ was consistently poor for the clustered representation across all cases. While Raw graphs maintain a more robust spatial fidelity ($R^2  \approx 0.90$  on seen data), the Clustered models often collapse, exhibiting $R^2$ values that frequently drop below $0$. On unseen architectures, this degrades further to significantly negative values (e.g., GCN: $-2.23$), indicating that clustered predictions are worse than a simple mean baseline.
    
 \textbf{The Representation Struggle:} Even with raw gate-level data, the Clock Skew MAE ($0.16$) remains significantly higher than the error for Wirelength ($0.06$) or Power ($0.06$). This suggests that while Raw graphs provide a more stable foundation than Clustered proxies, neither representation currently capture skew robustly. This suggests that future clustering algorithms in GNN EDA tasks must explicitly optimize for task alignment, ensuring that the specific attributes required for the downstream objective are mathematically preserved in the compressed representation.

\subsection{The Generalization Gap} \label{sec:generalization}
A critical finding of CTS-Bench is the contrast between In-Distribution and Out-of-Distribution test set performance. We explicitly selected the Zipdiv architecture as our hold-out set because it is fundamentally different from the training designs, as it is much smaller and has a distinct structural profile. On seen architectures, the models perform well. In particular, Raw GNNs preserve strong spatial accuracy, reaching an $R^2$ of about $0.90$. However, performance drops sharply on the unseen Zipdiv design. The Clustered models fail completely, producing negative $R^2$ values, which indicates that their predictions are statistically worse than a simple mean baseline. Even the Raw models struggle to generalize, with $R^2$ values only in the $0.0\text{--}0.2$ range. While this avoids the total collapse seen with clustering, it is still far from usable in practice.

These results imply that the failure on Zipdiv may not stem solely from modeling limitations, but could also reflect a lack of structural diversity within the training data. Consequently, future benchmarks might benefit from prioritizing broader regime diversity with designs varying significantly in scale and topology to better evaluate zero-shot generalization.


\subsection{Usage Scenarios}
\label{sec:use}

\noindent \textbf{Benchmarking next-generation GNN Accelerators:} Existing GNN benchmarks\cite{JMLR:v24:22-0567}\cite{NEURIPS2023_39f8ef62}, often based on social , citation networks, or molecular graph, fail to capture the structural characteristics of logic circuits. CTS-Bench addresses this by offering Raw Gate-Level graphs to challenge memory capacity and Clustered graphs to evaluate compute throughput. This facilitates the evaluation of Memory Efficiency on massive industrial netlists and Sparse Compute Support for the irregular connectivity of real-world circuits.

\vspace{0.5em}
\noindent \textbf{Graph Coarsening Testbench:} CTS-Bench quantifies the predictive fidelity loss associated with graph reduction. Although our baseline clustering yields a $13.3\times$ compression ratio, it compromises local clock skew accuracy. This highlights a key challenge: \textit{achieving high compression without sacrificing physical fidelity.} The benchmark enables the evaluation of different clustering methods in relevance to EDA tasks.

\section{Conclusion and Future Work}
\label{sec:conlcude}

In this paper, we presented CTS-Bench, a benchmark designed to quantify the impact of graph resolution on the performance of machine-learning-based EDA surrogates for Clock Tree Synthesis. Our experimental results demonstrate a clear accuracy–efficiency trade-off: while physics-aware graph clustering reduces peak memory usage by up to $17.2\times$, it consistently fails to preserve the fine-grained structural information required for reliable clock skew modeling. This degradation in predictive fidelity highlights a fundamental limitation of generic graph coarsening approaches and underscores the need to better balance scalability and accuracy in graph learning for ML-driven EDA tasks.  More broadly, this work serves as a starting point toward a comprehensive, CTS-centric benchmarking infrastructure for graph learning in physical design. Future extensions of CTS-Bench will incorporate additional placement constraints and sources of structural variation, including multi-\textit{V\textsubscript{t}} designs, macro-dominated layouts, and non-uniform power grid structures, to further reflect the complexity of modern industrial design flows.

Future work will address this limitation along three complementary directions. First, we will extend CTS-Bench to support task-aware and learned graph coarsening methods that optimize clustering for CTS-relevant objectives such as skew and power, mitigating the accuracy loss observed with generic clustering. Second, we will use CTS-Bench to study tighter integration between placement and Clock Tree Synthesis by training fast proxy models that provide clock-aware feedback during placement, enabling earlier optimization of clocking constraints. Finally, we will expand the benchmark to include a wider range of design architectures to evaluate zero-shot generalization across unseen designs and topologies.

\bibliographystyle{ACM-Reference-Format}
\bibliography{refs}
\end{document}